# Search-Based Fairness Testing: An Overview

*Hussaini Mamman
Department of Computer and Information Sciences
Universiti Teknologi PETRONAS
Seri Iskandar, Malaysia
hussaini_21000736@utp.edu.my

Shuib Basri
Department of Computer and Information Sciences
Universiti Teknologi PETRONAS
Seri Iskandar, Malaysia
shuib_basri@utp.edu.my

Abdullateef O. Balogun
Department of Computer and Information Sciences
Universiti Teknologi PETRONAS
Seri Iskandar, Malaysia
abdullateef.ob@utp.edu.my

Abdullahi Abubakar Imam
School of Digital Sciences
Universiti Brunei Darussalam
Brunei Darussalam, Brunei
abdullahi.imam@ubd.edu.bn

Ganesh Kumar
Department of Computer and Information Sciences
Universiti Teknologi PETRONAS
Seri Iskandar, Malaysia
ganesh_21000736@utp.edu.my

Luiz Fernando Capretz
Department of Electrical and Computer Engineering
Western University, London, Canada.
lcapretz@uwo.ca

*Abstract*—Artificial Intelligence (AI) has demonstrated remarkable capabilities in domains such as recruitment, finance, healthcare, and the judiciary. However, biases in AI systems raise ethical and societal concerns, emphasising the need for effective fairness testing methods. This paper reviews current research on fairness testing, particularly its application through search-based testing. Our analysis highlights progress and identifies areas of improvement in addressing AI systems' biases. Future research should focus on leveraging established search-based testing methodologies for fairness testing.

## I. INTRODUCTION

Artificial Intelligence (AI)-based systems have gained popularity over time. They are now integral components of many software systems, including those used for medical diagnoses, policing, loan approvals and risk assessments [1]. However, the rapid growth of AI-based systems in areas directly involving humans has raised concerns about their adoption's potential risks and challenges. One of the significant risks associated with AI is the possibility of discrimination and bias in their decision-making processes [2], [3].

A classic example comes from an AI-based system used by US courts to make pretrial detention and release decisions [4]. An examination of the system, Correctional Offender Management Profiling for Alternative Sanctions (COMPAS), revealed racial discrimination against black Americans [5]. Similarly, an AI system used to screen job applicants has been found to favour specific candidates over others based on gender [6]. In addition, minority homebuyers are reported to face widespread lending discrimination, causing 80% of Black mortgage applicants to be denied [7]. These cases where AI-based systems are found to be discriminatory are numerous, making it imperative to ensure that decisions made by AI-based systems are fair.

Fairness refers to the ability of AI-based systems to avoid biases and prevent discrimination [8]. The aim is to ensure that these systems produce fair outputs and behaviours for all inputs relevant to the task at hand, regardless of sensitive attributes like gender, race, or age. Specifically, the software should not discriminate against certain groups or individuals based on their characteristics [9]. Fairness has become a crucial requirement for AI-based systems to ensure their trustworthiness and widespread adoption [10], [11], making it imperative to test AI-based systems for fairness.

Fairness in AI-based systems cannot be guaranteed solely by developing better machine learning (ML) algorithms [2], as the discrimination can originate from various sources [12]. These include biases in the training dataset or the algorithm, and hyperparameters used to train the ML model [13].

Software testing is essential to software development, ensuring reliability and quality, and fairness testing can help detect and fix fairness issues in AI-based systems [3]. Fairness testing is a branch of software testing that evaluates how an AI-based system treats individuals or groups fairly and without discrimination [14]. Fairness testing aims to uncover as many discriminatory inputs as possible in an AI-based system.

Various software testing methods have been applied for fairness testing [15]. Combinatorial testing focuses on input interactions but can face combinatorial explosion [16]. Verification-based testing relies on constraint-solving techniques but is less scalable and resource-intensive [17]. A promising approach is search-based testing (SBT), which efficiently explores input spaces to detect challenging bugs [14], [18], [19]. For this, many fairness testing approaches employed SBT techniques to detect discrimination in AI-based systems. This study provides an overview of the fairness testing approaches that used SBT.

The rest of the paper is organised as follows. Section II gives a background of AI-based systems, fairness testing, and search-based fairness testing (SBFT). Section III highlights the methodology of the study, while results and discussions are provided in Section IV. Section V highlights some potential research directions. Section VI gives a concluding remark for the paper, while section VII finally displays the references used in the study.

## II. BACKGROUND

This section introduces AI-based systems, fairness testing and fairness testing life cycle. SBFT and its workflow are also discussed.

### A. AI-based Systems

AI-based (or just AI) systems are software applications with at least one AI component to provide functionalities. AI-based systems include various software tools and applications that use machine learning (ML) techniques to analyse data and develop analytical models. ML is a subfield of AI that focuses on data analysis and provides valuable insights to improve the performance of AI-based systems. Applying ML techniques enhances AI-based systems' accuracy, efficiency, and reliability[20].

AI-based systems include systems such as hiring decision support systems and healthcare systems that are used to diagnose patients. Image or speech recognition and autonomous driving are other examples [21].

*B. Fairness in AI-Based System*

Fairness in decision-making is the absence of bias or preference toward a person or group based on their inherent or acquired attributes [9]. Two notable definitions of fairness have been employed in fairness literature: individual fairness and group fairness. Individual fairness ensures that any two people who are similar in terms of a similarity metric defined for a particular task should get the same result [22]. For example, a loan software would be considered fair when it grants a loan to two individuals with similar characteristics, irrespective of their protected attributes. Individual fairness is more frequently addressed in fairness testing [22].

On the other hand, Group fairness involves ensuring that the distribution of outputs is similar across different groups based on a particular input characteristic. This approach seeks fairness by ensuring that the same proportion of individuals within various groups receives a specific outcome [2]. For example, in group fairness, a loan software would be considered fair with respect to age if it approves loans for the same proportion of applicants under 40 and over 40 years of age.

*C. Fairness Testing*

Fairness testing is crucial in the software engineering process as it ensures high-quality and reliable AI-based systems are developed and deployed. Fairness testing involves executing test cases to uncover discrimination in AI-based systems.

Fairness testing aims to assess the level of fairness a classifier provides by automatically generating test instances and using them to identify possible instances of discrimination [23]. Input instances that exhibit bias towards individuals with similar characteristics are called discriminatory instances [24]. These instances highlight the presence of unintended discrimination within AI systems and are a focus of concern in individual fairness testing and analysis.

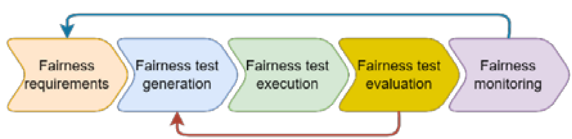

Fig. 1: Fairness testing life cycle [15]

*1) Fairness Testing Life Cycle*

The process of fairness testing involves a series of testing activities that detail its implementation [14]. As depicted in Fig. 1, software engineers determine and specify the desired fairness requirements for the AI-based system under test (SUT) via requirements engineering. Then, they identify and construct test oracles in alignment with these fairness criteria and generate or sample test inputs from the available data. The engineers then execute the test inputs on the SUT to determine if the test oracles are satisfied. They assess the tests' efficacy in revealing fairness bugs and use the bug report generated from the test run results to correct and eliminate such bugs. This fairness testing procedure is repeated until an acceptable level of fairness is attained for the SUT. When the system is deployed to production, continuous monitoring is applied to ensure that fairness requirements are satisfied.

*D. Search-Based Fairness Testing (SBFT)*

Search-based testing (SBT) methods typically involve systematically exploring the input space of a system to identify specific inputs that trigger certain behaviours or properties. SBT combines automatic test case generation and search techniques to optimise software testing [25]. A test case is an input of variables or conditions a tester uses to confirm that the SUT functions appropriately and meets a given requirement under review [25]. SBT can reduce the time and effort required to create test cases while increasing their effectiveness. It can also enable identifying more defects in the software in less time by selecting the best possible test cases to ensure the software is thoroughly tested [18].

Search-based fairness testing (SBFT) uses SBT techniques to evaluate the degree of discrimination in AI-based systems. It involves employing search algorithms and optimisation techniques to systematically explore the input space of a model and identify potential instances where unfair outcomes occur. The goal is to uncover and address discrimination cases in AI systems that may disproportionately affect specific individuals or groups based on race, gender, or other sensitive characteristics.

III. METHODOLOGY

In this section, we outline the methodology adopted for conducting the review. We initiated a thorough review of relevant papers published in the ACM & IEEE journals from 2017 to 2023.

*A. Keyword Search*

We intend to capture emerging trends within the realms of fairness testing research. The search criteria were devised using the keywords in Fig. 2 to ensure a more comprehensive search.

```
("bias" OR "discriminat*" OR "fair*") AND
("test*" OR "detect*" OR "audit" OR "evaluat*" OR "assess*"
OR "verif*" OR "discover*" OR "uncover*" OR "Investigat*" OR
"Examin*" OR "Inspect*" OR "check*") AND
("learn*" OR "software" OR "Artificial Intelligence" OR "AI" OR
"system*" OR "application" OR "Natural language processing"
OR "NLP" OR "Neural networks" OR "Algorithm" OR "Data
mining" OR "computer vision" OR "big data" OR "data-driven"
OR "decision making")
```

Fig. 2: Database search keywords

*B. The Review Process*

Fig. 3 illustrates the utilised flowchart in the research, which was adapted from Preferred Reporting Items for Systematic Reviews and Meta-Analyses (PRISMA) [26]. After removing duplicate papers and screening based on title and abstract, our review commenced with 53 articles. By applying our predetermined exclusion criteria, 33 articles were eliminated from consideration. As a result, this review is composed of a total of 20 articles. The exclusion criteria used in this study are as follows:

- Irrelevant to fairness testing in machine learning
- Absence of SBT methods for fairness evaluation

- Lack of emphasis on fairness test generation
- Duplicate publications
- Non-English papers

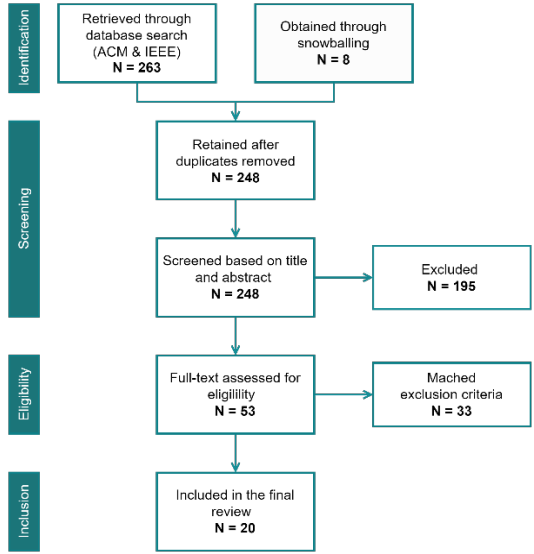

Fig. 3: Study PRISMA diagram

## C. Data Extraction

A data extraction form is designed to gather information from the selected articles on SBFT. The form has categories encompassing the author, publication year, research problem, core techniques, ML access level, ML task, data type, and evaluation metrics. The extracted data underwent thematic analysis to facilitate evaluation and interpretation, enabling insights from the data to form the research questions.

## IV. RESULT

This section showcases the results derived from the conducted review. The results are structured according to carefully formulated research questions, providing a guided framework for presenting the findings. The research questions are defined and discussed as follows.

*RQ 1: How has the field of search-based fairness testing evolved over the years?*

Fairness testing, specifically SBFT, is still a growing research area. In 2018, one article was published indicating the foundational stage. The subsequent years, 2019 and 2020, witnessed a consistent rise, with two articles showcasing growing interest and maturity. The trend continued in 2021 when five articles were published. Notably, 2022 experienced a significant surge with eight articles, highlighting a substantial boost in research activity. The trend continues in 2023 (until August), with three published articles indicating a dynamic, expanding research area with a growing impact. This trend is depicted in Fig. 4.

*RQ 2: What are the techniques employed for search-based fairness testing?*

An increasing number of fairness testing techniques used SBT to examine the input space of the AI-based system under test. These methods mainly detect individual discrimination and are based on ML classification tasks. For example, AEQUITAS [27] generates random inputs to find discriminatory instances and then applies perturbation to the non-protected attributes of those instances, using probabilistic search to discover neighbouring discriminatory samples. CGFT [16] enhances AEQUITAS by replacing the random search with combinatorial testing to generate a diverse test suite. Similarly, KOSEI [28] replaces the probabilistic search of AEQUITAS with sequential perturbation. SG [29] uses local explainability and symbolic execution to identify and generate discriminatory inputs based on the decision boundaries of SUT. Then, these inputs are perturbed for more discriminatory inputs. RULER [30] simultaneously perturbs sensitive and non-sensitive attributes to identify additional discriminatory instances outside the strict causal relations.

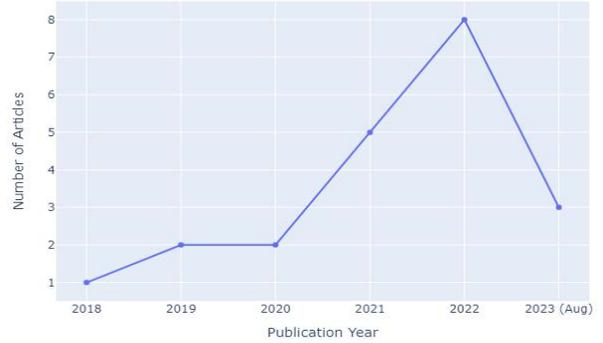

Fig. 4: Distribution of published studies over the years

ExpGA [31] employs local interpretability to identify seed instances that could induce discrimination upon flipping protected attributes, efficiently utilising these seeds to generate many discriminatory offspring through a genetic algorithm (GA). Xie and Wu [32] employed reinforcement learning (RL) to create test inputs for fairness testing by treating the ML model under test (MUT) as part of the RL environment. The RL agent alters the environment to produce discriminatory inputs, gauges the environment's state, and offers feedback through rewards. This iteration uncovered optimal strategies for generating effective discriminatory inputs.

In deep learning models, ADF [33], [34] uses gradient-guided search to generate discriminatory instances. Diverse seed instances are selected from clustered samples, and the area around each identified discriminatory instance is harnessed to generate more instances. EIDIG [35] improves ADF by introducing momentum terms to aid escape from local optima for higher success in detecting individual discriminatory instances and utilising prior gradient knowledge to optimise the generation of more discriminatory instances. NeuronFair [36] identifies biased neurons through analysis, generating instances to boost their Activation Difference (ActDif) values and further expanding the set by perturbing identified instances through nearby seed searches. DeepFAIT [37] utilises Generative Adversarial Networks (GANs) to transform images across sensitive domains and identifies fairness-related neurons through ActDif. Then, it generates test samples using image processing strategies.

Some methods address specific issues or different types of bias in addition to generating discriminatory instances. For example, Ma et al. [24] constructed initial discriminatory instances for fairness testing using an interpretable method. In contrast, LIMI [38] centred on creating test inputs based on naturalness. LIMI uses GANs to mimic the target model's decision boundary in a latent space, approximating data distribution with a surrogate linear limit and identifying potential discriminatory instances closer to the actual decision boundary through vector manipulations and calculations. The

*fAux* [39] method is designed to identify historical bias [12], and it achieves this goal by comparing the derivatives of the prediction model with those of an auxiliary model. The auxiliary model estimates the protected variable based on observed data, which helps avoid the need to generate counterexamples and prevents the inclusion of out-of-distribution inputs.

Distinct research focuses on various ML tasks. For instance, in regression-based tasks, Perera et al. [19] employed GA to generate test cases that could potentially unveil biases by assessing the maximum difference between predictions, referred to as the fairness degree. FairRec [40], [41] adopted a dual-particle swarm technique to gauge the maximum gap between user groups in recommendation tasks. This method employs separate swarms: one aimed at the most advantaged group and the other at the most disadvantaged group within the multidimensional search space. In addition, other studies focus on detecting group discrimination. For example, TestSGD [42] uses a rule-based method to measure group discrimination. It applies slight, uniform changes to a randomly sampled input to generate more samples. Then, the samples are used to estimate the statistical parity score between demographic groups. FAIRVIS [43] provides an interactive visual analytics tool that facilitates fairness auditing of SUT by integrating subgroup discovery techniques and performance comparison, aiding detailed investigation through coordinated views.

*RQ 3: Which fairness categories are examined by search-based fairness testing studies?*

From the work of Galhotra et al. [2] in 2017, the domain of SBFT has been primarily characterised by a focus on individual fairness testing. This emphasis is driven mainly by the claim that testing for individual discrimination leads to identifying more discriminatory instances [2]. As can be seen in Fig. 5, Group fairness testing in SBFT has received comparatively less attention (10%), with only two studies [41], [42] dedicated to its exploration.

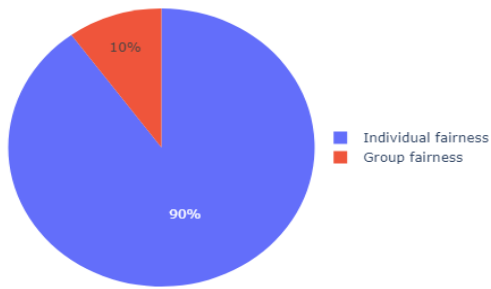

Figure 5: Studies distribution based on fairness category

*RQ 4: Which fairness evaluation metrics are used in search-based fairness testing studies?*

Numerous assessment metrics have been devised in fairness testing, grounded in the foundational concepts of individual and group fairness. When evaluating individual discrimination, research predominantly employs three benchmarks: 1) the quantity of generated test cases, 2) the proportion of detected discriminatory instances, and 3) the time required for test case generation. In studies concentrating on group fairness, the Calders & Verwer (CV) score is commonly utilised to quantify the disparity in misclassification rates between the two evaluated groups.

## V. RECOMMENDATION FOR FUTURE STUDIES

This section highlights potential research directions that need further exploration in the field of SBFT as follows.

*1) Using metaheuristic algorithms*

SBT techniques systematically explore a system's input space to uncover specific inputs that trigger desired behaviours, often employing established metaheuristic algorithms like genetic GA, particle swarm optimisation (PSO), and simulated annealing. These algorithms use an objective function to guide an efficient search for optimal solutions from a large search space. While a few studies [19], [31], [40], [41] have used GA and PSO for fairness testing, there's a need for more exploration of other well-known metaheuristics in assessing AI systems for fairness.

*2) Multi-objectives testing*

Existing SBFT studies focus on generating test cases that reveal bias. However, ensuring fairness in AI systems often involves balancing multiple objectives, such as fairness metrics, accuracy, and interpretability [44]. Multi-objective optimisation techniques seek to find a set of solutions that represent a trade-off between multiple competing objectives. In fairness testing, this could mean generating inputs that simultaneously expose biases while maintaining model performance, an area that needs further investigation.

*3) Initial seed selection*

The efficiency of generating discriminatory instances is affected by the quality of initial seeds [33], [35]. However, existing studies used random or clustering-based sampling, which may not be optimal. Notably, only one study [24] has explored initial seed selection in fairness testing, highlighting a potential avenue for future research.

*4) Re-using test data*

Re-using test data is essential in software testing for efficiency and quality, but in studies related to SBFT, generating new test data for each protected attribute can lead to redundant cycles and unnecessary waste of time and effort. The possibility of using test cases designed for one attribute to assess another emphasises the need for efficient strategies like test case recycling, including sound pruning [2], to optimise resource utilisation.

*5) Test cost reduction*

Methods for reducing costs, such as test selection, prioritisation, and minimisation, have been explored in SBT [45]. High test costs hinder fairness testing as it involves model retraining, repeated predictions, or extensive data generation. Exploring cost-reduction techniques for more efficient fairness testing is required.

*6) More fairness testing options*

Most SBFT research has concentrated on classification tasks, with limited exploration of regression and recommendation tasks. Other ML tasks like reinforcement learning and unsupervised learning remain under-explored, offering potential for future investigation. Additionally, while existing SGFT methods prioritise individual fairness, group fairness, especially subtle discrimination [42], is less studied. Novel testing strategies for group fairness are needed.

## VI. CONCLUSIONS

Ensuring the fairness of AI-based systems is of utmost importance to uphold justice and equity in society. As AI

continues to expand, the need for fairness testing remains a critical area of research. In this context, search-based testing methods will continue to have a significant role. This paper reviews contemporary fairness testing approaches utilising SBT to evaluate AI-based systems. Additionally, we have outlined potential avenues for future research, offering valuable insights for further exploration.